\documentclass[11pt,a4paper]{article}
\usepackage[hyperref]{emnlp2020}
\usepackage{times}
\usepackage{latexsym}

\usepackage{microtype}

\usepackage{url}
\usepackage{nicefrac} 
\usepackage{booktabs} 
\usepackage{amsfonts} 
\usepackage{graphicx}
\usepackage{amsmath}
\usepackage{algorithm}
\usepackage{algorithmic}
\usepackage{multirow}
\usepackage{hyperref}
\usepackage{caption}

\usepackage{amsmath,amsfonts,bm}

\newcommand\und[1]{\underline{#1}}

\def\eqref#1{equation~\ref{#1}}

\def\1{\bm{1}}

\def\vw{{\bm{w}}}

\def\mS{{\bm{S}}}

\DeclareMathAlphabet{\mathsfit}{\encodingdefault}{\sfdefault}{m}{sl}
\SetMathAlphabet{\mathsfit}{bold}{\encodingdefault}{\sfdefault}{bx}{n}

\DeclareMathOperator*{\argmax}{arg\,max}
\DeclareMathOperator*{\argmin}{arg\,min}

\aclfinalcopy 
\def\aclpaperid{247} 

\title{Neural Mask Generator: Learning to Generate\\
       Adaptive Word Maskings for Language Model Adaptation}

\author{Minki Kang$^{1*}$ \quad\quad Moonsu Han$^1$\thanks{\quad Equal contribution.} \quad\quad Sung Ju Hwang$^1$$^,$$^2$ \\  \\
	KAIST$^1$, Daejeon, South Korea \\ AITRICS$^2$, Seoul, South Korea \\
	\texttt{\{zzxc1133, mshan92, sjhwang82\}@kaist.ac.kr}}

\date{}

\begin{document}
\maketitle

\setlength{\abovedisplayskip}{3pt}  
\setlength{\belowdisplayskip}{3pt}
\setlength{\belowcaptionskip}{-7.5pt}
\captionsetup[table]{skip=1pt}

\begin{abstract}
We propose a method to automatically generate a domain- and task-adaptive maskings of the given text for self-supervised pre-training, such that we can effectively adapt the language model to a particular target task (e.g. question answering). Specifically, we present a novel reinforcement learning-based framework which learns the masking policy, such that using the generated masks for further pre-training of the target language model helps improve task performance on unseen texts. We use off-policy actor-critic with entropy regularization and experience replay for reinforcement learning, and propose a Transformer-based policy network that can consider the relative importance of words in a given text. We validate our \emph{Neural Mask Generator (NMG)} on several question answering and text classification datasets using BERT and DistilBERT as the language models, on which it outperforms rule-based masking strategies, by automatically learning optimal adaptive maskings.
\footnote{Code is available at \url{github.com/Nardien/NMG}.}
\end{abstract}
\section{Introduction}
The recent success of the \emph{language model pre-training} approaches~\citep{BERT, ELMo, GPT2, T5, XLNet}, which train language models on diverse text corpora with self-supervised or multi-task learning, have brought up huge performance improvements on several natural language understanding (NLU) tasks~\citep{GLUE, SQuAD}. The key to this success is their ability to learn generalizable text embeddings that achieve near optimal performance on diverse tasks with only a few additional steps of fine-tuning on each downstream task.

Most of the existing works on language model aim to obtain a universal language model that can address nearly the entire set of available natural language tasks on heterogeneous domains. Although this train-once and use-anywhere approach has been shown to be helpful for various natural language tasks~\citep{BERT, GPT2, UnifiedLM, T5}, there have been considerable needs on adapting the learned language models to domain-specific corpora (e.g. healthcare or legal). Such domains may contain new entities that are not included in the common text corpora, and may contain only a small amount of labeled data as obtaining annotation on them may require expert knowledge. 
Some recent works~\citep{HowtoFT, BioBERT, SciBERT, DontStopPT} suggest to further pre-train the language model with self-supervised tasks on the domain-specific text corpus for adaptation, and show that it yields improved performance on tasks from the target domain.

Masked Language Models (MLMs) objective in BERT~\citep{BERT} has shown to be effective for the language model to learn the knowledge of the language in a bi-directional manner~\citep{Transformer}.
In general, masks in MLMs are sampled at random~\citep{BERT, RoBERTa}, which seems reasonable for learning a generic language model pre-trained from scratch, since it needs to learn about as many words in the vocabulary as possible in diverse contexts.

However, in the case of further pre-training of the already pre-trained language model, such a conventional selection method may lead a domain adaptation in an inefficient way, since not all words will be equally important for the target task. Repeatedly learning for uninformative instances thus will be wasteful. Instead, as done with instance selection~\cite{Specialist, MentorNet, DVRL, L2TL}, it will be more effective if the masks focus on the most important words for the target domain, and for the specific NLU task at hands. How can we then \emph{obtain} such a masking strategy to train the MLMs? 

Several works~\citep{SpanBERT, ERNIE, ERNIE2, SpanSelection} propose rule-based masking strategies which work better than random masking ~\citep{BERT} when applied to language model pre-training from scratch. Based on those works, we assume that adaptation of the pre-trained language model can be improved via a \emph{learned} masking policy which selects the words to mask. Yet, existing models are inevitably suboptimal since they do not consider the target domain and the task. To overcome this limitation, in this work, we propose to adaptively generate mask by learning the optimal masking policy for the given task, for the task-adaptive pre-training~\citep{DontStopPT} of the language model.

\begin{figure}
    \centering
    \includegraphics[width=0.95\linewidth]{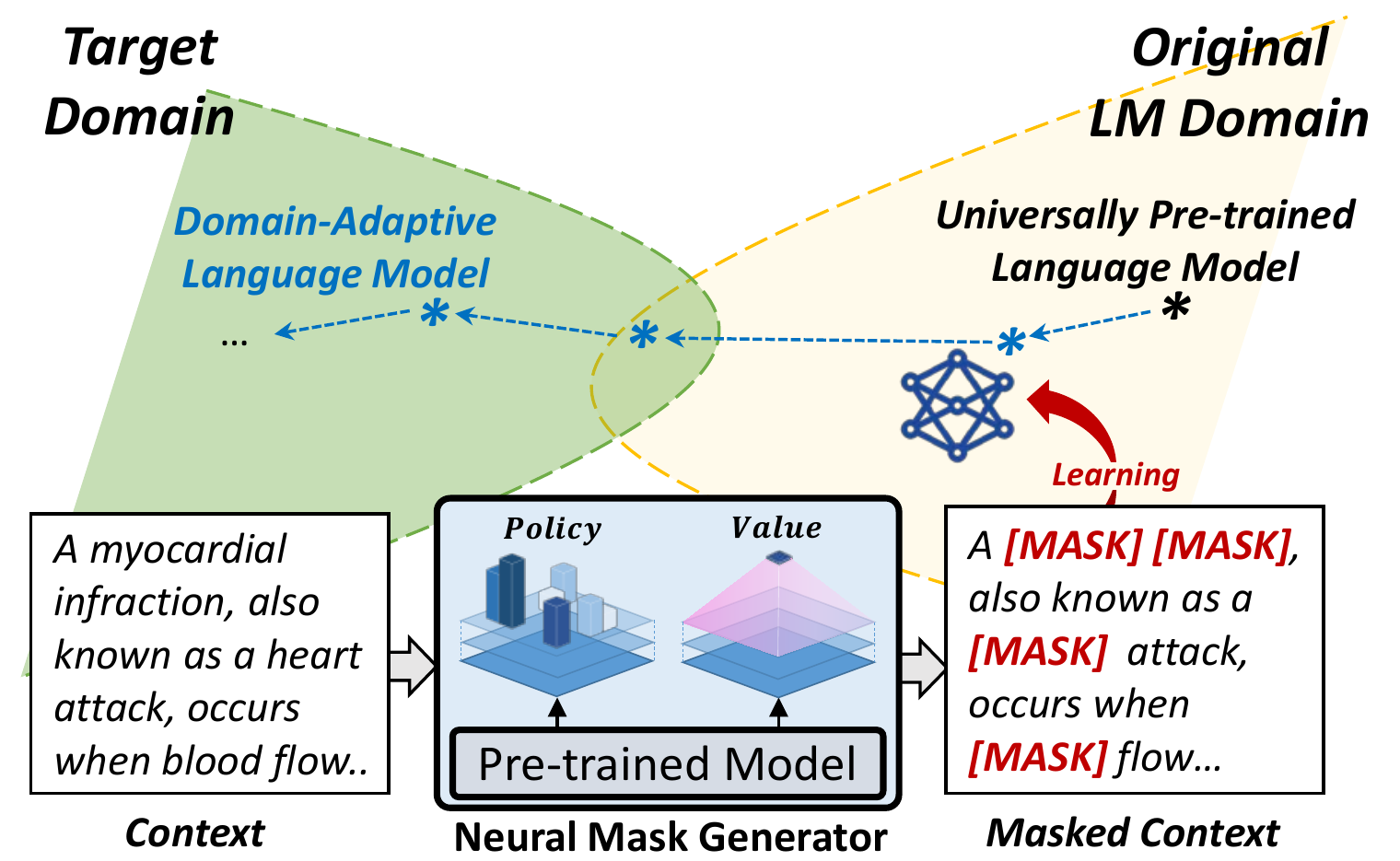}
    \vskip -0.1in
    \caption{\small \textbf{Concept.} Pre-training on domain text leads the language model parameters to adapt to the given target domain. We assume that adjusting the masking policy of the MLM objective affects the training trajectory of the language model, such that it moves towards a better solution space for the target domain. 
    This illustration of the solution spaces for the two domains is motivated by \cite{DontStopPT}.}
    \label{fig:concept_figure}
\end{figure}

As described in Figure \ref{fig:concept_figure}, we want to further pre-train the language model on a specific task with a task-dependent masking policy, such that it directs the solution to the set of parameters that can better adapt to the target domain, while task-agnostic random policy leads the model to an arbitrary solution.  To tackle this problem, we pose the given learning problem as a meta-learning problem where we learn the task-adaptive mask-generating policy, such that the model learned with the masking strategy obtains high accuracy on the target task.  We refer to this meta-learner as the \textbf{N}eural \textbf{M}ask \textbf{G}enerator (NMG). Specifically, we formulate mask learning as a bi-level problem where we pre-train and fine-tune a target language model in the inner loop, and learn the NMG at the outer loop, and solve it using renforcement learning. We validate our method on diverse NLU tasks, including question answering and text classification. The results show that the models trained using our NMG outperforms the models pre-trained using rule-based masking strategies, as well as finds a proper adaptive masking strategy for each domain and task.

Our contribution is threefold:
\begin{itemize}
	\item We propose to learn the mask generating policy for further pre-training of masked language models, to obtain optimal maskings that focus on the most important words for the given text domain and the NLU task. 
	\item We formulate the problem of learning the task-adaptive mask generating policy as a bi-level meta-learning framework which learns the LM in the inner loop, and the mask generator at the outer loop using reinforcement learning. 
    \item We validate our mask generator on diverse tasks across various domains, and show that it outperforms heuristic masking strategies by learning an optimal task-adaptive masking for each LM and domain. We also perform empirical studies on various heuristic masking strategies on the language model adaptation.
\end{itemize}

\section{Related Work}

\paragraph{Language Model Pre-training} 
Ever since \citet{UniLM} suggested language model pre-training with multi-task learning, inspired by the success of fine-tuning on ImageNet pre-trained models on computer vision tasks~\citep{Transfer}, research on the representation learning for natural language understanding tasks~\nocite{Word2Vec} have focused on obtaining a global language model that can generalize to any NLU tasks. A popular approach is to use self-supervised pre-training tasks for learning the contextualized embedding from large unannotated text corpora using auto-regressive~\citep{ELMo, XLNet} or auto-encoding~\citep{BERT, RoBERTa} language modeling. Following the success of the Masked Language Model (MLM) from~\citep{BERT, RoBERTa}, several works have proposed different model architecture~\citep{ALBERT, T5, ELECTRA} and pre-training objectives~\citep{ERNIE, ERNIE2, RoBERTa, SpanBERT, SpanSelection, UnifiedLM}, to improve upon its performance. Some works have also proposed alternative masking policies for the MLM pre-training over random sampling, such as SpanBERT~\cite{SpanBERT} and ERNIE~\cite{ERNIE}. Yet, none of the existing approaches have tried to \emph{learn} the task-adaptive mask in a context-dependent manner which is the problem we target in this work. 

\paragraph{Language Model Adaptation}
Pre-training the language model on the target domain, then fine-tuning on downstream tasks, is the most simple yet successful approach for adapting the language model to a specific task. Some studies~\citep{BioBERT, SciBERT, DATBERT} have shown the advantage of further pre-training the language model on a large unlabeled text corpus collected from a specific domain. Moreover, \citet{HowtoFT} and \citet{ unsuperivsed-domain-adaptation} investigate the effectiveness of further pre-training of the language model on small domain-related text corpora. Recently, \citet{DontStopPT} integrates prior works and defines domain-adaptive pre-training and task-adaptive pre-training, showing that domain adaptation of the language model can be done with additional pre-training with the MLM objective on a domain-related text corpus, as well as a smaller but directly task-relevant text corpus.

\paragraph{Meta-Learning}
Meta-learning~\cite{Learningtolearn} aims to train the model to generalize over a distribution of tasks, such that it can generalize to an unseen task. There exist large number of different approaches to train the meta-learner~\citep{MANN, MatchingNetwork, OptimML, MAML, DARTS}.
However, existing meta-learning approaches do not scale well to the training of large models such as masked language models. Thus, instead of the existing meta-learning method such a gradient based approach, we formulate the problem as a bi-level problem~\cite{Bilevel} of learning the language model in the inner loop and the mask at the outer loop, and solve it using reinforcement learning. Such optimization of the outer objective using RL is similar to the formulation used in previous works on neural architecture search~\cite{Reinforce-NAS, Reinforce-NAS2}.
\section{Problem Statement}
\label{section:Problem_formulation}
We now describe how to formulate the problem of \emph{learning to generate masks} for the Masked Language Model (MLM) as a bi-level optimization problem. Then, we describe how we can reformulate it as a reinforcement learning problem.

\subsection{Masked Language Model}
For pre-training of the language models, we need an unannotated text corpus $\mS = \{ s^{(1)}, \cdots, s^{(T)} \}$. Here the $s = [w_1, w_2, \cdots, w_N]$ is the context, whose element $\vw_i\in{s}$ is a single word or token. To formulate the meta-learning objective, we assume each context corpus as the part of the task $\mathcal{T} = \{ \mS, D_{tr}, D_{te} \}$ consisting of the context and its corresponding task dataset. For the MLM, we need to generate a noisy version of $s$, which we denote as $\hat{s}$. Let $z_i$ be the indicator of masking $i$-th word of $s$. If $z_i = 1$, $i$-th word is replaced with the [MASK] token. The objective of the MLM then is to predict the original token of each [MASK] token. Therefore, we can formulate this problem as follows:

\vspace{-0.4cm}
\begin{equation*} 
\min_{\theta} \mathcal{L}_{MLM}(\theta;\mS,\mathbf{\hat{S}}) = \sum_{s,\hat{s} \in \mS,\mathbf{\hat{S}}} -\log p_\theta (\bar{s} | \hat{s})
\end{equation*}

\noindent where $\theta$ is the parameter of the language model and $\bar{s}$ is the original tokens of each [MASK] token, and $\mathbf{\hat{S}}$ is the set of masked contexts. Following the formulation from \citep{XLNet}, we can approximate the MLM objective with $z_i$ as follows:

\vspace{-0.5cm}
\begin{equation}
\label{eqn:approx}
\max_{\theta} \log p_\theta (\bar{s} | \hat{s}) \approx \sum_{i=1}^N z_i \log p_\theta (w_i | \hat{s})
\end{equation}
\begin{equation*} 
p_\theta (w_i | \hat{s}) = \frac{\exp (H_\theta(\hat{s})^\top_i e(w_i))}{\sum_{w'} \exp (H_\theta(\hat{s})_i^\top e(w'))}
\end{equation*}

\noindent where $H_\theta (\hat{s})$ indicates the contextualized representation of $\hat{s}$ from the language model layer (e.g. Pre-trained Transformer layers in \citep{BERT}), and $e(w_i)$ denotes the embedding of word $w_i$ from the last prediction layer. Our objective then is to learn the optimal policy for determining each mask indicator $z_i$, which we will describe in detail in the next subsection.

\subsection{Bi-level formulation}
\label{section:metalearning}
We now describe how to formulate this learning problem as a bi-level problem consisting of inner and outer level objectives. Consider that $z_i$ can be represented using an arbitrary function $\mathcal{F}$ parameterized with $\lambda$:

\vspace{-0.6cm}
\begin{equation}
\label{eqn:policy}
\pi_\lambda(a_t = i | s) = \mathcal{F}_\lambda(w_i)
\end{equation}

\vspace{-0.4cm}
\begin{equation}
\label{eqn:action}
\mathbf{a_\lambda} = \argmax_{a} \prod_{t} \pi_\lambda(a_t | s)
\end{equation}

\vspace{-0.2cm}
\begin{equation}
\label{eqn:mask}
z_{\lambda, i} =
\begin{cases}
1, & \mbox{if }i \in \mathbf{a_\lambda} \\
0, & \mbox{if }i \notin \mathbf{a_\lambda}
\end{cases}
\end{equation}

\noindent where $\pi_\lambda(a_t = i | s)$ is the probability of masking i-th word $w_i$ from $s$, and $a_\lambda$ indicates the list of word indices to be masked and $z_{\lambda, i}$ is the mask indicator parameterized by the parameter $\lambda$. The details of corresponding equations will be described in section~\ref{section:rl}. Therefore, the MLM objective has been slightly changed from its original form, into the following objective:

\vspace{-0.3cm}
\begin{align}
\begin{split}
\label{eqn:MLM_objective}
\theta(\lambda){}& = \argmin_{\theta} \mathcal{L}_{MLM}(\theta, \lambda;\mS,\mathbf{\hat{S}}) \\
                 & = \argmax_{\theta} \sum_{s \in \mS} \log p_\theta (\bar{s} | \hat{s}) \\
                 & \approx \argmax_{\theta} \sum_{s \in \mS} \sum_{i=1}^N z_{\lambda, i} \log p_\theta (w_i | \hat{s})
\end{split}
\end{align}

\noindent where the masked context $\hat{s}$ is parameterized by the parameter $\lambda$. Now assume that we have found the optimal parameter $\theta(\lambda)$ for language model from equation \ref{eqn:MLM_objective}. Then, we need to fine-tune the language model on the downstream task. 
Although linear heads used in both pre-training and fine-tuning are different, we describe the parameter of both models as $\theta(\lambda)$ for simplicity. Following the bi-level framework notation described in \citep{Bilevel}, the inner objective function for fine-tuning can be written as follows:

\vspace{-0.3cm}
\begin{equation}
\label{eqn:finetune_objective}
\min_{\theta(\lambda)}\mathcal{L}_{train}(\theta(\lambda), \lambda) = \sum_{(x,y) \in D_{tr}} l(g_{\theta(\lambda)}(x), y)
\end{equation}

\noindent where $D_{tr}$ is a training dataset, $l$ is the loss function of the supervised learning, and $g_{\theta(\lambda)}$ is the function representation of downstream task solver model. In case of question answering task, each $x$ consists of a context and corresponding question and $y$ is the corresponding answer spans.

Assume that we find optimal parameter $\theta^{*}(\lambda)$ from supervised task fine-tuning. Then, the final outer-level objective can be described as follows:

\vspace{-0.3cm}
\begin{align}
\begin{split}
\label{eqn:outer_objective}
\min_{\lambda} {}& \mathcal{L}_{test}(\lambda, \theta^*(\lambda)) = \sum_{(x,y) \in D_{te}} l(g_{\theta^{*}(\lambda)}(x), y) \\
                 & s.t. \quad \theta^*(\lambda) = \argmin_{\theta(\lambda)} \mathcal{L}_{train}(\theta(\lambda), \lambda) \\
                 & \ and \quad \theta(\lambda) = \argmin_\theta \mathcal{L}_{MLM}(\theta, \lambda)
\end{split}
\end{align}

\noindent This will allow us to obtain the optimal parameter $\lambda^{*}$ which minimizes the task objective function on a test dataset $D_{te}$. 

Although the outer objective is differentiable, we formulate the optimization problem of the outer objective as a reinforcement learning problem to avoid excessive computation cost caused by the two constraint terms. 

\paragraph{Justification of Reinforcement Learning}
In this paragraph, we explain why we use the Reinforcement Learning (RL) instead of the differentiable method to train the parameter $\lambda$.
As indicated in the equation~\ref{eqn:outer_objective}, our inner loop includes consecutive two steps of language model training. The NMG model $\lambda$ is addressed to the pre-training step rather than the task fine-tuning step. Therefore, the direct differentiation of the outer objective contains two second-derivative terms for both the MLM loss $\mathcal{L}_{MLM}$ and the task train loss $\mathcal{L}_{train}$. With a single-step approximation, the derivative of the outer objective is approximated as follows:
\begin{align*}
    &\ \nabla_{\lambda} \mathcal{L}_{test} (\lambda, \theta^*(\lambda)) \\
    &\approx \nabla_\lambda \mathcal{L}_{test} (\lambda, \tilde{\theta}^*(\lambda)) \\
    &- \alpha \nabla^2_{\lambda, \theta} \mathcal{L}_{MLM} (\theta, \lambda) \nabla_\theta \mathcal{L}_{test} (\tilde{\theta}^*(\lambda), \lambda) \\
    &- \beta \nabla^2_{\lambda, \tilde{\theta}(\lambda)} \mathcal{L}_{train} (\tilde{\theta}(\lambda), \lambda) \nabla_{\tilde{\theta}(\lambda)} \mathcal{L}_{test}(\tilde{\theta}(\lambda), \lambda)
\end{align*}

\noindent where $\tilde{\theta}(\lambda) = \theta - \alpha \nabla_\theta \mathcal{L}_{MLM} (\theta, \lambda), \tilde{\theta}^*(\lambda) = \tilde{\theta}(\lambda) - \beta \nabla_{\tilde{\theta}(\lambda)} \mathcal{L}_{train} (\tilde{\theta}(\lambda), \lambda)$ are approximated parameters, and $\alpha, \beta$ are learning rates.
Such gradient estimation requires high computational costs since it includes the computation of Hessian-product vectors of the massive language model's parameters approximated as 110 millions~\citep{MAML, BERT}.

Instead, we can address the first-order approximation ($\alpha=0, \beta=0$) to the derivative of the outer objective to avoid second-order derivative computation as follows:
\begin{equation*}
    \nabla_\lambda \mathcal{L}_{test}(\lambda, \theta^*(\lambda)) \approx \nabla_\lambda \mathcal{L}_{test}(\theta)
\end{equation*}
\noindent where $\theta^*(\lambda)$ is approximated to $\theta$.
Such approximation trivially results in a meaningless optimization since it ignores the pre-training step induced by the parameter $\lambda$, which decides the masking policy~\citep{DARTS}. Therefore, we approach solving this optimization problem with RL instead of the differential method to avoid such an issue. In the next section, we introduce how we formulate this problem as the RL with the outer objective as a non-differentiable reward.

\subsection{Reinforcement learning formulation}
\label{section:rl}
We now propose a reinforcement learning (RL) framework, which given the context $s$ as the state, decides on the actions $\mathbf{a} = \{ a_1, \cdots, a_T \}$ where $T$ is the number of masked tokens in the given context, each $a_t \in [1, N]$ is the token index that indicates a decision on masking the token $w_{a_t}$ following equation~\ref{eqn:mask}, and $a_1 \ne a_2 \ne \cdots \ne a_T$. The objective of the RL agent then is to find an optimal masking policy that minimizes $E(\lambda, \theta)$ from the section~\ref{section:metalearning}. In addition, minimizing $E(\lambda, \theta)$ on $D_{te}$ can be seen as maximizing its performance. Therefore, the objective is the maximization of the performance of the model on $D_{te}$. We can induce it by setting the reward $R$ as the accuracy improvement on $D_{te}$. We will describe the detail of the reward design in section \ref{section:NMG}.

In general RL formulation~\citep{RL} following Markov Decision Process (MDP), state transition probability can be described as $p(s_{t+1} | s_t, a_t)$. The probability of masking $T$ tokens is formularized as $p(\hat{s} | s) = \prod_{t=1}^{T} p(s_{t+1}|s_t, a_t)$, where $\hat{s}$ consists of $T$ number of [MASK] tokens. Although the representation of $s_t$ and $s_{t+1}$ are slightly different because of the addition of [MASK] token, we can approximate them as $s_t \approx s_{t+1}$ following the approximation in equation \ref{eqn:approx}, which inductively approximates representations after each word masking as a representation of original context.

Therefore, we can approximate the probability of masking $T$ tokens from the MDP problem to the problem without state-transition formulation as follows:

\vspace{-0.3cm}

\begin{equation*}
p(\hat{s} | s) = \prod_{t=1}^{T} p(s_{t+1}|s_t, a_t) \approx \prod_{t=1}^{T} \pi(a_t | s) 
\end{equation*}

where $a_t$ denotes the index of masked word included in the context $s$ and $\pi(\cdot | s)$ is the masking policy of the agent. By this approximation, we do not need to consider the trajectory along the temporal horizon, which makes the problem much easier. Instead, we approximate the problem as the task of selecting multiple discrete actions simultaneously for the same state.
We approximate the policy with neural network parameterized by $\lambda$ as $\pi_\lambda(a|s)$. As in equations~\ref{eqn:action} and ~\ref{eqn:mask}, the mask $z_{\lambda, i}$ is determined by actions generated from the neural policy $\pi_\lambda(a|s)$. In the next section, we will describe how to train the Neural Mask Generator to generate the neural policy $\pi_\lambda(a|s)$ maximizing the reward $R$ using RL in Section \ref{section:NMG}.
\section{Neural Mask Generator}
\label{section:NMG}
\begin{figure}[t]
    \centering
    \includegraphics[width=1\linewidth]{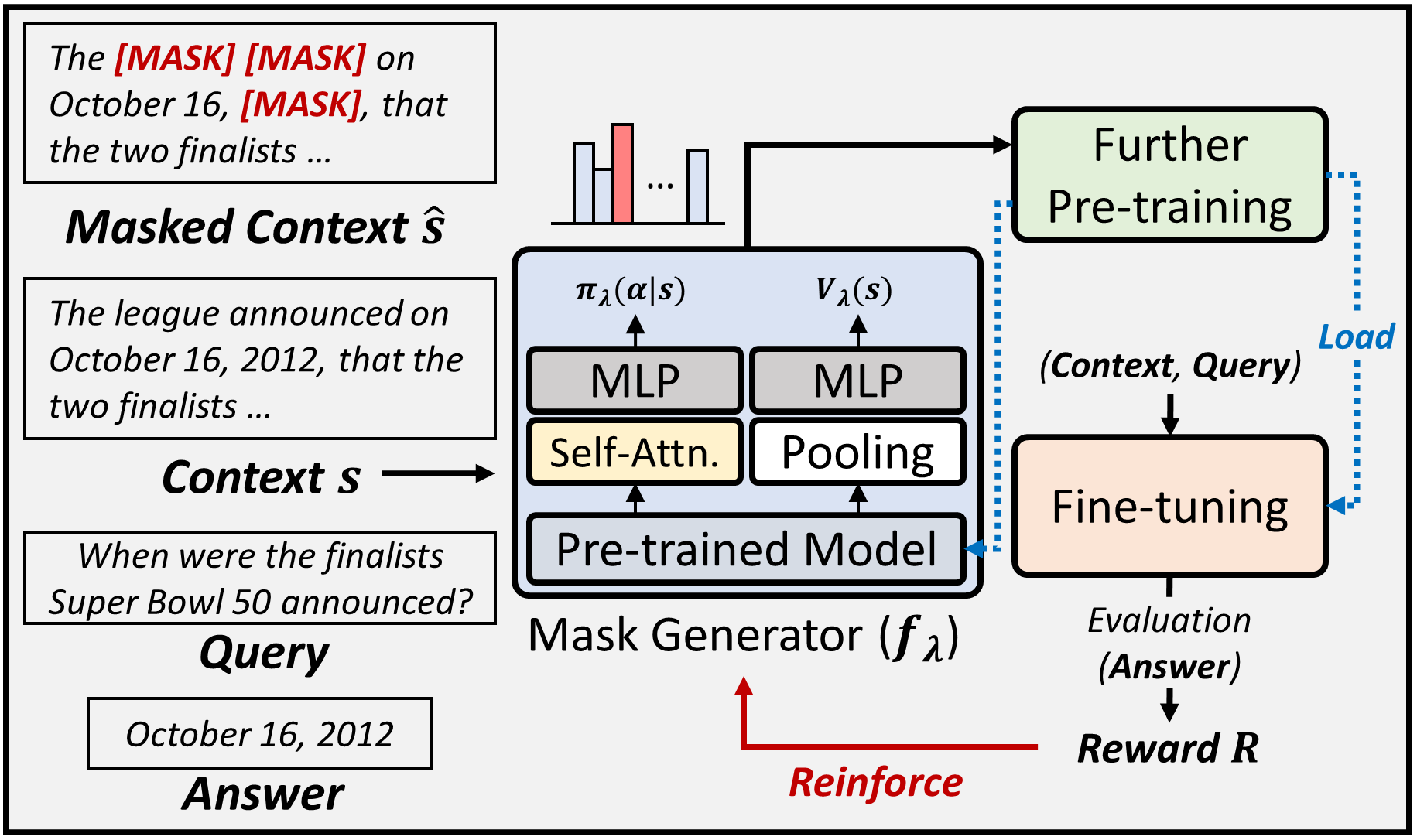}
    \vspace{-0.8cm}
    \caption{\small The overview of the meta-training framework for our Neural Mask Generator (NMG). In each episode, masked contexts by the NMG are used for further pre-training. Then, the further pre-trained language model is used in both the mask generator and fine-tuning.}
    \label{fig:fullarchitecture}
\end{figure}

In this section, we describe our model, \emph{Neural Mask Generator (NMG)}, which learns the masking policy to generate an optimal mask on an unseen context using deep RL with the detailed descriptions of the 
framework setup.
The overview of the meta mask generator framework is shown in Figure~\ref{fig:fullarchitecture}. For detailed descriptions of the approaches, the procedures of both training and test phase, and algorithm, please \textbf{see Appendix~\ref{sec:algorithm}}.

\subsection{Reinforcement Learning Details}

\paragraph{Model Architecture}
The probability of selecting $i$-th word $w_i$ of $s$ as $t$-th action $a_t$ can be described as $\pi_\lambda(a_t = i | s)$, where $ \sum_i \pi_\lambda(a_t = i | s) = 1 $. Instead of $\mathcal{F}_\lambda$ in equation~\ref{eqn:policy}, the neural policy from the NMG is given as follows:

\begin{equation*}
\pi_\lambda(a_t = i | s) = \frac {\exp(f_\lambda (H_{\theta'}(s)_i))}{\sum_{t} \exp(f_\lambda (H_{\theta'}(s)_t))}
\end{equation*}

\noindent where $f_\lambda$ is a deep neural network parameterized by $\lambda$, which has a self-attention layer~\citep{Transformer} followed by linear layers with gelu activation~\cite{GELU}, and $H_{\theta'}(s)_i$ is the contextualized representation of $w_i$ of context $s$ from the frozen language model layer $\theta'$. 
Note that $\theta'$ is shared with the target language model $\theta$ and not trained during the NMG training. Further, $f_\lambda (H_{\theta'}(s)_i)$ outputs the scalar logit of the $w_i$, and the final probabilistic policy is computed by the softmax function.

\paragraph{Training Objective}

We train the NMG model using the Advantage Actor-Critic method~\citep{A3C} with the value estimator. Furthermore, we resort to off-policy learning~\citep{Off-PolicyAC} such that the agent can explore the optimal policy based on its past experiences. To this end, we leverage a prioritized experience replay, in which we store every state, action, reward, and old-policy pairs~\cite{DQN}, and sample them based on their absolute value of the advantage~\cite{PER}. We use importance sampling to estimate the value of the target policy $\pi_\lambda$ with the samples from the behavior policy $\pi_{\lambda_{old}}$, which are sampled from the replay buffer~\citep{Off-PolicyAC, TRPO, PPO, ACER}.
To sum up, the objectives for both policy and value network are as follows:
\begin{align}
\begin{split}
\label{eqn:PolicyObjective}
\mathcal{L}_{policy} =  \sum_{(s,a,R,\pi_{\lambda_{old}})} {}& - \frac{\pi_\lambda(a | s)}{\pi_{\lambda_{old}}(a | s)} (R - V_\lambda(s)) \\
                                         & - \alpha \ \mathcal{H}(\pi_\lambda(a | s))
\end{split}
\end{align}

\begin{equation}
\label{eqn:ValueObjective}
\mathcal{L}_{value} = \sum_{(s,a,R,\pi_{\lambda_{old}})} \frac{1}{2} (R - V_\lambda(s))^2
\end{equation}

\noindent where the $(s, a, R, \pi_{\lambda_{old}})$ is set of sampled replays from the replay buffer, $\mathcal{H}$ is an entropy function $\mathcal{H}(\pi_\lambda(a|s)) = -\sum_t \pi_\lambda(a_t|s) \log \pi_\lambda(a_t|s)$, and $\alpha$ is a hyperparameter for entropy regularization. $V_\lambda(s)$ is an estimated value of the state $s$. The value network consists of linear layers with activation function after mean pooling $\frac{1}{N} \sum_{t} H_{\theta'}(s)_t$. 

To summarize, the outer-level objective for updating the NMG parameter $\lambda$ can be written as follows:
\begin{equation}
\label{eqn:rl}
    \min_\lambda \mathcal{L}_{policy} + \mathcal{L}_{value}
\end{equation}
At each episode of meta-training, we update the NMG parameter $\lambda$ by optimizing the above objective function.

\paragraph{Reward Design and Self-Play}
As in Section~\ref{section:rl}, the reward function is considered as the accuracy improvement on the test set $D_{te}$. Therefore, the pre-training step (equation \ref{eqn:MLM_objective}) and the fine-tuning then evaluation step (equation \ref{eqn:finetune_objective}, \ref{eqn:outer_objective}) should be done to get the reward in every episodes\textbf{}.

Since using the full size of dataset in the inner loop is generally not feasible, we randomly sample smaller sub-task $\mathcal{T}' = \{ \mS', D_{tr}', D_{val} \}$ from $\mathcal{T}$ at every episode. For the evaluation, we randomly split the training set $D_{tr}$ to generate a hold-out validation set $D_{val}$ and replace $D_{te}$ in equation \ref{eqn:outer_objective} to $D_{val}$ while meta-training where $D_{te}$ is unobservable.
We use a sufficiently large hold-out validation set $D_{val}$ to prevent the masking policy from overfitting to  $D_{val}$.
We assume that the meta-learner NMG also performs well on $\mathcal{T}$ if it is trained on diverse sub-tasks $\mathcal{T}'$ where $|\mathcal{T}| \gg |\mathcal{T}'|$.

The problem to be considered for using diverse sub-tasks is that the NMG model encounters different sub-task $\mathcal{T}'$ at every new episode. Since $\mathcal{T}'$ determines the state distribution $\mS'$ and the data $D_{tr}$ to be trained, it results in the reward scale problem that the expectation of the validation accuracy on $D_{val}$ varies depending on the composition of $\mathcal{T}'$ then makes it harder to evaluate the performance increment of the neural policy across episodes.

To address this problem, we introduce the random policy as an opponent policy to evaluate the neural policy relative to it. Therefore, the reward R is defined as $sgn(r-b)$, where $sgn$ is the sign function, $r$ and $b$ are the accuracy score on $D_{val}$ from neural and random policy respectively. However, the random policy may be too weak as the opponent. To overcome this limitation, we add another neural policy as an additional opponent to induce the zero-sum game of two learning agent by the concept of the self-play algorithm~\citep{SelfPlay2, SelfPlay1}.

Then, three distinct policies are compared with each other during episodes and two neural policies are individually trained.
Furthermore, in each neural agent training, only actions corresponding to disjoint comparing with others are stored in a global replay buffer for a more accurate reward assignment of each action. 

\paragraph{Continual Adaptive Learning}
\label{section:continual}
For fair comparison at every episode, the same language model should be used to evaluate the policy. 
Initializing the language model before the start of each inner loop can be the simplest choice to handle this.
However, since we pre-train the language model for only few steps during each episode of meta-training, the model is always evaluated around the original language model domain (see Figure \ref{fig:concept_figure}). To avoid this, at each episode, the language model which is pre-trained by the NMG model of former step is continually loaded instead of the fixed checkpoint except for the first episode. By this, we intend that our agent learns the optimal policy of various environments. Furthermore, the agent can learn dynamic policy based on the learned degree of the target language model.
\section{Experiment}
We now experimentally validate our Neural Mask Generator (NMG) model on multiple NLU tasks, including question answering and text classification tasks using two different language models, and analyze its behaviors. In Section~\ref{sec:results}, we evaluate the NMG with several baselines. Then, we evaluate the effect of the specific design choices made for our model through ablation studies in Section~\ref{sec:ablation}. Finally, we analyze how the policy learned by the NMG model works in Section~\ref{sec:analysis}.

\begin{table*}[t]
\centering
\small
\begin{tabular}{clcccccc}
\toprule
                                                &        & \multicolumn{2}{c}{SQuAD} & \multicolumn{2}{c}{emrQA} & \multicolumn{2}{c}{NewsQA} \\
Base LM                                         & Model  & EM          & F1          & EM          & F1          & EM           & F1          \\
\midrule
\multirow{7}{*}{BERT}                           & No PT  & $80.63_{0.32}$      & $88.18_{0.25}$       & $70.58_{0.09}$      & $76.60_{0.29}$      & $\und{51.66_{0.31}}$& $\und{66.23_{0.16}}$\\
                                                & Random & $80.64_{0.02}$      & $88.14_{0.10}$       & $71.40_{1.01}$      & $77.31_{0.78}$      & $51.46_{0.19}$   & $66.21_{0.22}$\\
                                                & Whole  & $80.73_{0.23}$      & $88.20_{0.11}$       & $\und{71.65_{0.33}}$& $\bf77.84_{0.31}$   & $51.12_{0.28}$      & $65.94_{0.51}$\\
                                                & Span   & $80.66_{0.19}$      & $88.19_{0.11}$       & $70.87_{0.38}$      & $76.76_{0.33}$      & $51.17_{0.63}$      & $65.93_{0.50}$\\
                                                & Entity & $\und{80.79_{0.13}}$& $\und{88.23_{0.22}}$ & $71.50_{0.34}$      & $\und{77.57_{0.27}}$& $51.55_{0.43}$  & $65.44_{0.34}$\\
                                                & Punct. & $80.60_{0.24}$      & $88.07_{0.25}$       & $71.17_{0.58}$      & $77.08_{0.57}$      & $51.47_{0.37}$      & $66.18_{0.36}$\\ \cmidrule{2-8} 
                                                & NMG    & $\bf{80.83_{0.20}}$ & $ \bf88.28_{0.21}$   & $\bf71.84_{0.68}$   & $77.49_{0.55}$      & $\bf{51.81_{0.33}}$& $\bf{66.57_{0.48}}$   \\ 
\midrule
\multicolumn{1}{l}{\multirow{7}{*}{DistilBERT}} & No PT  &$76.75_{0.41}$&$85.13_{0.26}$&$68.52_{0.39}$      &$75.00_{0.53}$      &$\und{48.61_{0.39}}$&$\und{63.45_{0.56}}$ \\
\multicolumn{1}{l}{}                            & Random &$76.46_{0.35}$      &$84.92_{0.17}$      &$69.02_{0.40}$      &$75.70_{0.29}$      &$48.52_{0.46}$&$63.06_{0.21}$ \\
\multicolumn{1}{l}{}                            & Whole  &$76.48_{0.27}$      &$84.96_{0.15}$      &$\und{69.64_{0.39}}$  &$\und{76.16_{0.24}}$  &$48.22_{0.41}$&$62.92_{0.33}$ \\
\multicolumn{1}{l}{}                            & Span   &$76.73_{0.13}$      &$84.96_{0.13}$      &$69.54_{0.21}$&$76.11_{0.33}$&$48.59_{0.09}$ & $63.15_{0.39}$ \\
\multicolumn{1}{l}{}                            & Entity &$76.34_{0.36}$      &$84.78_{0.21}$      &$69.25_{0.43}$      &$75.98_{0.39}$      &$48.19_{0.20}$ & $62.97_{0.16}$ \\
\multicolumn{1}{l}{}                            & Punct. &$\und{76.85_{0.09}}$ &$\und{85.16_{0.03}}$      &$69.41_{0.21}$      &$76.14_{0.18}$                 &$48.58_{0.26}$ & $63.14_{0.20}$ \\\cmidrule{2-8} 
\multicolumn{1}{l}{}                            & NMG  & $\bf 76.93_{0.32}$   &$\bf 85.30_{0.23}$  &$\bf69.98_{0.37}$      &$\bf76.51_{0.45}$      &$\bf 48.75_{0.08}$ & $\bf 63.55_{0.14}$   \\ 
\bottomrule
\end{tabular}
\caption{\small Performance of various masking strategies on the QA tasks. We run the model three times with different random seeds then report the average performances, with standard deviations (subscripts). The numbers in bold fonts denote best scores, and the numbers with underlines denote the second best scores.}
\label{table:1}
\end{table*}

\paragraph{Tasks and Datasets}
For question answering, we use three datasets, namely SQuAD v1.1~\cite{SQuAD}, NewsQA~\cite{NewsQA}, and emrQA ~\cite{emrQA} to validate our model. We use the MRQA\footnote{https://mrqa.github.io/} version for both SQuAD and NewsQA to sustain a coherency. 
We also preprocess emrQA to fit the format of other datasets. We use a standard evaluation metric named Exact-Matchs (EM) and F1 score for question answering task. For text classification, we use IMDb~\cite{IMDb} and ChemProt~\cite{Chemprot}, following the experimental settings of \citep{DontStopPT}.

\paragraph {Baselines}
According to \citet{BERT} and \citet{SpanBERT}, training the language models with difficult objectives is much more beneficial when pre-training from scratch. To test whether it is also the case for task-adaptive pre-training, we experiment with two heuristic masking strategies, which we refer to as whole-random and span-random. In addition, we also tested the named entity masking proposed by~\citet{ERNIE} (entity-random). Below is the complete list of the heuristic baselines we compare against in the experiments.

\textbf{1) No-PT}
A baseline without any further pre-training of the language model.

\textbf{2) Random}
A random masking strategy introduced in BERT~\cite{BERT}.

\textbf{3) Whole-Random}
A random masking strategy which masks the entire word instead of the token (sub-word). This method is introduced by the authors of BERT~\cite{BERT}\footnote{https://github.com/google-research/bert}.

\textbf{4) Span-Random}
A random masking strategy which selects multiple consecutive tokens.

\textbf{5) Entity-Random}
A random masking strategy which selects named entities with highest priorities, then randomly selects other tokens.

\textbf{6) Punctuation-Random}
A random masking strategy which selects punctuation tokens first, then randomly selects other tokens.

\paragraph{Implementation Details}

For the language model $\theta$, we use the same hyperparameters and architecture with DistilBERT~\cite{DistilBERT} model (66M params) and BERT$_{BASE}$~\cite{BERT} model (110M params). Our implementation is based on the huggingface's Pytorch implementation version~\citep{HuggingFacesTS, PyTorch}.
We load the pre-trained parameters from the checkpoint of each language model in meta-testing and the first episode of meta-training.
As for the text corpus $\mS$ to pre-train the language model, we use the collection of contexts from the given NLU task.
We only use the Masked Language Model (MLM) objective for further pre-training. In the initial stage of meta-training, the NMG randomly selects actions for exploration. In meta-testing, it takes maximum probability indices as actions. 
We describe the details of language model training in \textbf{Appendix~\ref{sec:hyperparameters}} since we use different settings for each task and experiments. As for reinforcement learning (RL), we use the off-policy actor-critic method described in Section \ref{section:NMG}. For more details of the reinforcement learning framework, please \textbf{see Appendix~\ref{sec:hyperparameters}}.
\nocite{AdamW}\nocite{Adam}

\subsection{Results}
\label{sec:results}
First of all, we need to discuss how the MLM works on the language model pre-training. In practice, the Cloze task~\cite{Cloze} benefits when words that need to learn are masked. However, for the MLM, we observed that the masking prevents learning the masked words in the language model. Rather the MLM learns representations and relations between non-masked words by predicting the masked words using them. In the case of adaptive pre-training, learning the domain-specific vocabulary is crucial for the domain adaptation.
Therefore, masking out trivial words (e.g. Punctuations) may be more beneficial than masking out unique words (e.g. Named Entities) for the language model to learn the knowledge of a new domain. 
To see how it works, we experiment for punctuation-random, which masks only the punctuations, which are clearly useless for the domain adaptation.

In Table~\ref{table:1} and~\ref{table:2}, we report the performance of baselines and our model on both question answering and text classification tasks.
From the baseline results of Table~\ref{table:1}, we speculate on the important aspects of a masking strategy for better adaptation. The whole-word, span and entity maskings often lead to better results since it makes the MLM objective more difficult and meaningful~\citep{SpanBERT, ERNIE}.
For instance, in emrQA, most words are tokenized to sub-words since contexts include a lot of unique words such as medical terminologies. Therefore, whole-word masking could be most suitable for domain adaptation on emrQA.
In contrast, such maskings sometimes lead to worse results than random masking on some domains such as in NewsQA. 
Especially, in the case of DistilBERT, punctuation masking performs better than others. These result suggests that masking complicated words is rather disturbing for the adaptation of small language model.

On the other hand, our NMG learns the optimal masking policy for a given task and domain in an adaptive manner on any language model. In Table~\ref{table:1} and \ref{table:2}, we can see that this adaptive characteristic of our model makes the neural masking results in better or at least comparable performances to the baselines for all tasks. We further analyze the learned masking strategy in Section~\ref{sec:analysis}.

\begin{table}[t]
\centering
\small
\begin{tabular}{clcc}
\toprule
\multicolumn{1}{l}{}  &        & ChemProt       & IMDb          \\
Base LM               & Model  & Acc            & Acc           \\ 
\midrule
\multirow{7}{*}{BERT} & No PT  & $80.40_{0.70}$ & $92.28_{0.05}$ \\
                      & Random & $\und{81.25_{0.72}}$ & $92.45_{0.21}$ \\
                      & Whole  & $80.18_{1.20}$ & $\bf{92.55_{0.04}}$ \\
                      & Span   & $78.06_{1.72}$ & $92.40_{0.10}$ \\
                      & Entity & $79.68_{1.32}$ & $92.38_{0.13}$ \\
                      & Punct. & $79.68_{0.30}$ & $92.40_{0.18}$ \\ \cmidrule{2-4} 
                      & NMG    & $\bf81.66_{0.37}$ & $\und{92.53_{0.03}}$\\
\bottomrule
\end{tabular}
\caption{\small Performance of various masking strategies on the text classification tasks. The presentation format is the same as in Table~\ref{table:1}.}
\label{table:2}
\vspace{-0.2cm}
\end{table}

\begin{figure*}[t]
    \centering
    \includegraphics[width=0.96\linewidth]{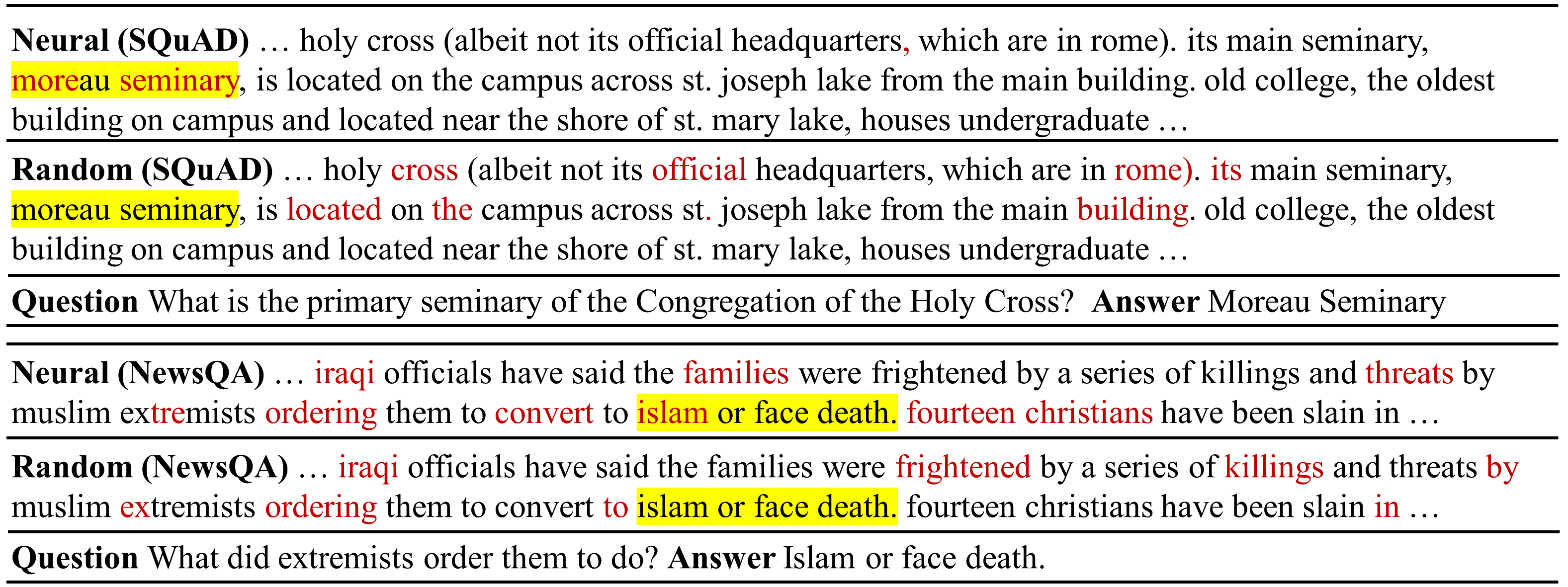}
    \vspace{-0.3cm}
    \caption{\small Examples of masked tokens using our Neural Mask Generator (Neural) model and random sampling (Random). The tokens colored in red denote masked tokens by the model, and words highlighted with a yellow box is the answer to the given question. For more examples on multiple datasets, \textbf{please see Appendix~\ref{sec:examples}.}}
    \label{fig:example}
\vspace{-0.2cm}
\end{figure*}

\begin{table}
\centering
\small
\resizebox{0.46\textwidth}{!}{
    \begin{tabular}{lcc}
    \toprule
    \small{Base LM:}            & \multicolumn{2}{c}{NewsQA} \\
    DistilBERT                  & EM          & F1          \\ 
    \midrule
    NMG                         & $\bf48.75_{0.08}$ & $\bf63.55_{0.14}$   \\
    NMG w/o Self-Play           & $48.64_{0.27}$ & $63.37_{0.17}$       \\
    NMG w/o Continual-Adaptive  & $48.52_{0.12}$ & $63.06_{0.20}$       \\ 
    \bottomrule
    \end{tabular}
}
\caption{\small Ablation Results on the NewsQA dataset.}
\label{table:3}
\end{table}

\subsection{Ablation study}
\label{sec:ablation}

\paragraph{Effectiveness of Self-Play}
We further investigate the effectiveness of self-play by comparing it with the NMG model without self-play, where the model only competes with the random agent. We validate this on the NewsQA dataset. The result in Table \ref{table:3} shows that the NMG model with self-play obtains better performance than its counterpart without self-play. This result verifies that competing with the opponent neural agent while learning helps the NMG model to learn better policy. 
\paragraph{Continual Adaptation}
We also perform an ablation study of the continual adaptation learning. 
The result in Table \ref{table:3} shows that the continual-adaptive masking strategy is significantly effective for the language model adaptation. The result suggests that helpful words for the language model to learn depends on the adaptation degree of it.

\subsection{Analysis}
\label{sec:analysis}

\begin{figure}[t!]
    \centering
    \includegraphics[width=0.9\linewidth]{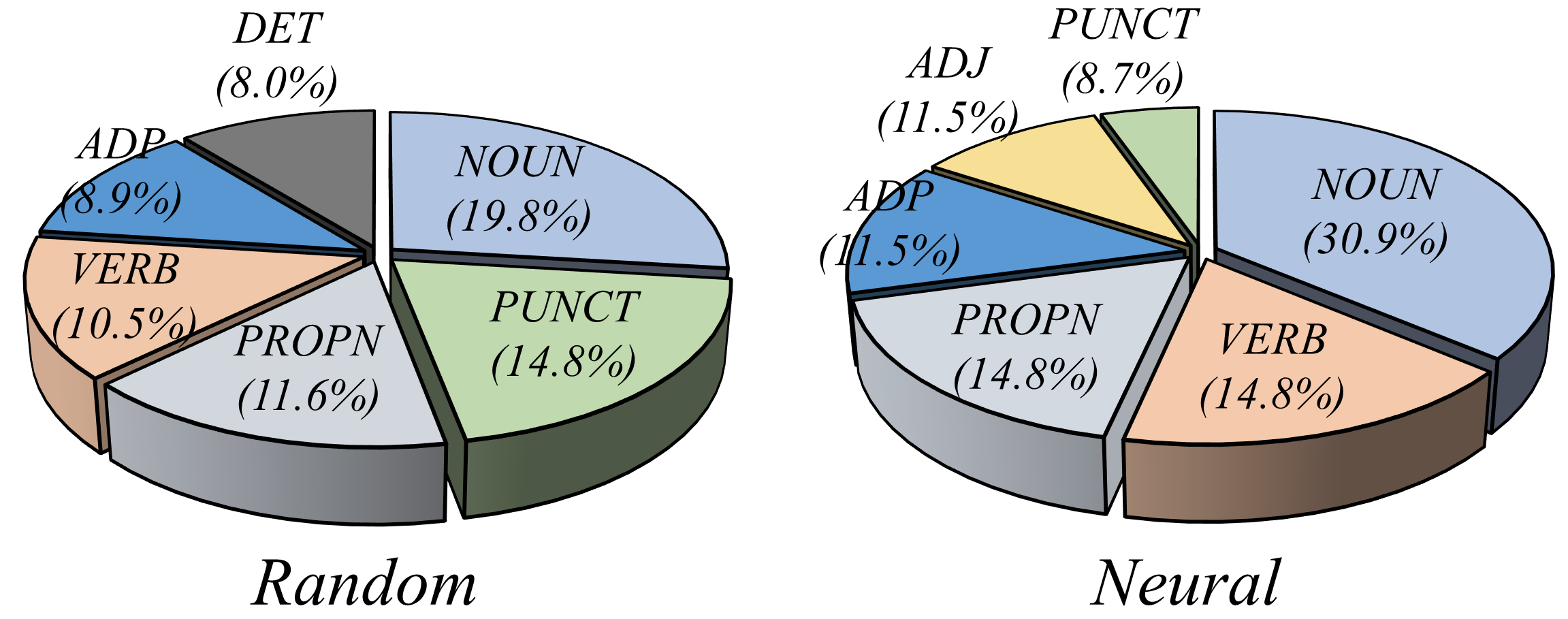}
    \vskip -0.1in
    \caption{\small Top-6 POS of Masked Words on NewsQA.}
    \label{fig:mask_statistics}
\end{figure}

\paragraph{Masked Word Statistics}
To analyze how our model performs, we measure the difference between which kind of word token is masked by both the random and neural policy on the pre-trained checkpoint. For qualitative analysis, we provide examples of masked tokens on the context in Figure~\ref{fig:example}. As shown in Figure~\ref{fig:example}, NMG tends to mask highly informative words such as \textit{seminary} or \textit{islam}, which are parts of the answer spans.
Furthermore, we analyze the masking behavior of our NMG by performing Part-of-Speech (POS) tagging on the masked words using spaCy\footnote{https://spacy.io}.  Figure~\ref{fig:mask_statistics} shows the six most frequent tags for the words masked out by the random and neural policy.
Figure~\ref{fig:mask_statistics} shows that the neural policy masks more words in noun, verb, and proper noun tags than the random policy, suggesting that our NMG model learns that masking such informative words is beneficial to adapt on the NewsQA task with the BERT$_{BASE}$ model as a language model.

\begin{figure}[t!]
    \centering
    \includegraphics[width=1.0\linewidth]{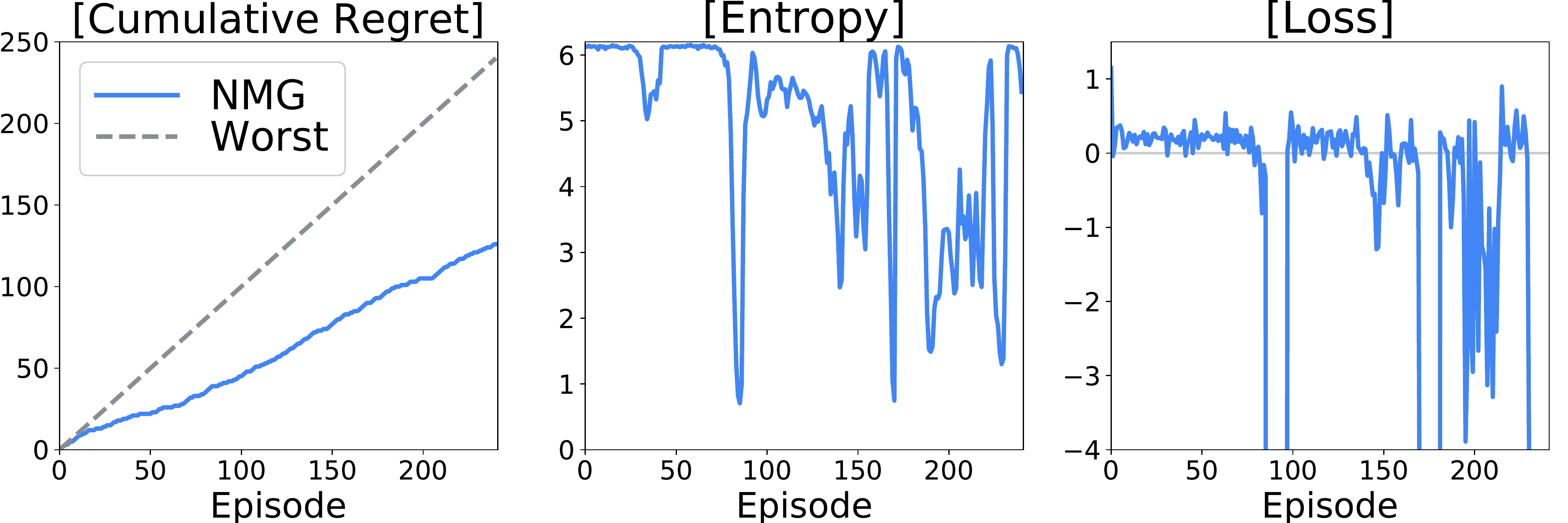}
    \vskip -0.1in
    \caption{\small Reinforcement learning curves on NewsQA.}
    \label{fig:learning_plot}
    \vskip -0.15in
\end{figure}

\paragraph{Learning Curves}
As already known, the RL-based methods often suffer from the instability problem. Therefore, we further analyze the learning curves of the NMG training in this section.
In Figure~\ref{fig:learning_plot}, we plot three kinds of learning curves to show the detailed training process.
\textit{Cumulative Regret} indicates how many times the neural agent is defeated against the random agent until certain episodes. The grey plot indicates the worst case that the random agent always defeats against the neural agent. \textit{Entropy} indicates the average entropy of policy for states given in a certain episode. Lower entropy means that the policy has a high probability of a few significant actions. \textit{Loss} indicates the RL loss described in the equation~\ref{eqn:rl}. We ignore outliers in the loss plot for brevity.

From the entropy and loss plots, we can notice that the policy converges as learning proceeds. 
However, it seems that such convergence is not continually sustained. From the cumulative regret plot, we can observe that the neural policy still often loses against the random policy, although it is trained for a while.
Such instability may come from the difficulty of the exact credit assignment on each action.
Otherwise, continuous change of state distribution from the continual adaptive learning may hinder the neural policy's convergence. 

Even if the NMG shows the notable results, there is room for improvement on RL in terms of efficiency and stability. We leave it as the future work.

\section{Conclusion}
We proposed a novel framework which automatically generates an adaptive masking for masked language models based on the given context, for language model adaptation to low-resource domains. To this end, we proposed the \emph{Neural Mask Generator (NMG)}, which is trained with reinforcement learning to mask out words that are helpful for domain adaptation. We performed an empirical study of various rule-based masking strategies on multiple datasets for question answering and text classification tasks, which shows that the optimal masking strategy depends on both the language model and the domain. We then validated NMG against rule-based masking strategies, and the results show that it either outperforms, or obtains comparable performance to the best heuristic. Further qualitative analysis suggests that such good performance comes from its ability to adaptively mask meaningful words for the given task.

\section*{Acknowledgments}
This work was supported by Samsung Advanced Institute of Technology (SAIT), the Engineering Research Center Program through the National Research Foundation of Korea (NRF) funded by the Korea Government MSIT (NRF2018R1A5A1059921), Institute for Information \& communications Technology Promotion(IITP) grant funded by the Korea Government (MSIT) (No.2016-0-00563, Research on Adaptive Machine Learning Technology Development for Intelligent Autonomous Digital Companion, and No.2019-0-00075, Artificial Intelligence Graduate School Program (KAIST)), and a study on the ``HPC Support" Project supported by the `Ministry of Science and ICT' and NIPA.

\bibliographystyle{acl_natbib}
\bibliography{emnlp2020}

\newpage
\appendix

\section{Algorithms}
\label{sec:algorithm}
We provide the pseudocode of algorithm for meta-training of the Neural Mask Generator (NMG).

In the case of meta-testing, \textit{InnerLoop} with the full task $\mathcal{T}$, pre-trained language model checkpoint $\theta$, and trained policy $\pi_\lambda$ as inputs can be considered as the meta-testing algorithm.

\renewcommand{\algorithmicrequire}{\textbf{Input:}}
\renewcommand{\algorithmicensure}{\textbf{Output:}}
\begin{algorithm}[h]
    \caption{NMG Training Algorithm}
\begin{algorithmic}
    \STATE Initialize random policy $\psi \sim \textit{Uniform}$
    \STATE Initialize two neural agents $\lambda, \lambda^{op}$ for Self-Play
    \STATE Initialize replay buffer $\mathcal{D}, \mathcal{D}^{op}$
    \STATE Arbitrarily split $D_{tr} \rightarrow D_{tr}, D_{val}$
    \STATE Load pre-trained language model $\theta$
    \WHILE {not done}
    \STATE Randomly Sample $\mathcal{T}'$ from $\mathcal{T}$
    \STATE {$r_\psi$, $\mathcal{E}_\psi$, $\theta(\psi) =$ \textit{InnerLoop}($\mathcal{T}', \theta, \psi$)}
    \STATE {$r_{op}$, $\mathcal{E}_{\lambda^{op}}$, $\theta(\lambda^{op}) =$ \textit{InnerLoop}($\mathcal{T}', \theta, \pi_{\lambda^{op}}$)}
    \STATE {$r$, $\mathcal{E}_{\lambda}$, $\theta(\lambda) =$ \textit{InnerLoop}($\mathcal{T}', \theta, \pi_{\lambda}$)}
    \STATE $\mathcal{A}^{op} \leftarrow \mathcal{E}_\psi, r_\psi, \mathcal{E}, r, \mathcal{E}_{op}, r_{op}$
    \STATE $\mathcal{D}^{op}, \lambda^{op} \leftarrow$ \textit{OuterLoop}($\mathcal{D}^{op}$, $\mathcal{A}^{op}$, $\lambda^{op}$)
    \STATE $\mathcal{A} \leftarrow \mathcal{E}_\psi, r_\psi, \mathcal{E}_{op}, r_{op}, \mathcal{E}, r$
    \STATE $\mathcal{D}, \lambda \leftarrow$ \textit{OuterLoop}($\mathcal{D}$, $\mathcal{A}$, $\lambda$)
    \STATE $\theta \leftarrow \theta(\lambda)$
    \ENDWHILE
\end{algorithmic}
\end{algorithm}

\begin{algorithm}[h]
    \caption{InnerLoop}
\begin{algorithmic}
    \STATE \textbf{Input:}
    \STATE $\quad$ Task $\mathcal{T}$, LM $\theta$, Policy $\pi$
    \STATE \textbf{Output:}
    \STATE $\quad$ Episode buffer $\mathcal{E}$, Accuracy $r$, LM $\theta(\lambda)$
    \STATE Initialize episode buffer $\mathcal{E}$
    \STATE $\mathbf{\hat{S}} = \{\}$, $\mS, D_{tr}, D_{val} \leftarrow \mathcal{T}$
    \FOR {$s$ in $\mS$}
    \STATE Masking $s$ following equation \ref{eqn:action}, \ref{eqn:mask}
    \STATE $\mathbf{\hat{S}} \leftarrow \mathbf{\hat{S}} \cup \hat{s}$, $\mathcal{E} \leftarrow \mathcal{E} \cup (s,a, \pi)$
    \ENDFOR 
    \STATE \textit{\# Actually, below two updates are done with mini-batch optimization with multiple steps}
    \STATE $\theta(\lambda) \leftarrow \theta - \nabla_\theta \mathcal{L}_{MLM}(\theta, \lambda; \mS, \mathbf{\hat{S}})$
    \STATE $\theta^*(\lambda) \leftarrow \theta(\lambda) - \nabla_{\theta(\lambda)} \mathcal{L}_{train}(\theta(\lambda), \lambda; D_{tr})$
    \STATE Evaluate $\theta$ on $D_{val}$ and acquire $r$
\end{algorithmic}
\end{algorithm}

\begin{algorithm}[t!]
    \caption{OuterLoop}
\begin{algorithmic}
    \STATE \textbf{Input:}
    \STATE $\quad$ $\mathcal{D}$, $\mathcal{E}_\psi$, $r_\psi$, $\mathcal{E}_{op}$, $r_{op}$, $\mathcal{E}$, $r$, Agent $\lambda$
    \STATE \textbf{Output:}
    \STATE $\quad$ Replay buffer $\mathcal{D}$, Trained Agent $\lambda$
    \STATE $R \leftarrow sgn(r-r_{op})$
    \FOR {$(s,a)$ in $\mathcal{E}$}
    \IF {$(s,a) \notin \mathcal{E} \cap \mathcal{E}_{op}$ and $(s,a) \notin \mathcal{E} \cap \mathcal{E}_{\psi}$}
    \STATE $R \leftarrow \min(sgn(r-r_\psi), sgn(r-r_{op}))$
    \ELSIF {$(s,a) \notin \mathcal{E} \cap \mathcal{E}_{op}$}
    \STATE $R \leftarrow sgn(r-r_{op})$
    \ELSIF {$(s,a) \notin \mathcal{E} \cap \mathcal{E}_{\psi}$}
    \STATE $R \leftarrow sgn(r-r_\psi)$
    \ELSE
    \STATE continue
    \ENDIF
    \STATE {$\mathcal{D} \leftarrow \mathcal{D} \cup \{(s, a, R, \pi_{old})\}$}
    \ENDFOR
    \STATE Sample a mini-batch $\{(s, a, R, \pi_{old})\}$ from $\mathcal{D}$
    \STATE $\lambda \leftarrow \lambda + \nabla_\lambda \mathcal{L}$ following equation \ref{eqn:PolicyObjective}, \ref{eqn:ValueObjective} using sampled replays
\end{algorithmic}
\end{algorithm}

\section{Hyperparameters}
\label{sec:hyperparameters}
\subsection{Reinforcement Learning (Outer Loop)}
We describe detailed hyperparameters in Table~\ref{table:rl-hyper} for reinforcement learning (RL). We use the prioritized experience replay~\cite{PER} with exponent value as 1. In addition, we address the concept of susampling~\citep{Word2Vec} on replay sampling. Specifically, we divide the priority of each replay by the square root of word frequency within the corresponding context. For optimization, we use Adam~\cite{Adam} optimizer to train the NMG model and its opponent (Self-Play).

\begin{table}[!ht]
\centering
\begin{tabular}{ll}
\toprule
\multicolumn{1}{c}{Hyperparameters} & \multicolumn{1}{c}{Value} \\ \midrule
Learning Rate                       & 0.0001                    \\
Number of Epochs                    & 10                        \\
Minibatch Size                      & 64                        \\
Replay Buffer Size                  & 50000                     \\
Entropy Regularization              & 0.01                      \\
Maximum Episodes                    & 200                       \\
\bottomrule
\end{tabular}
\caption{Hyperparameters for Reinforcement Learning Training (Outer Loop). QA is short for Question Answering and TC is short for Text Classification.}
\label{table:rl-hyper}
\end{table}

\begin{table*}[!ht]
\caption{Hyperparameters for LM Training of Meta-Train (Inner Loop).}
\label{table:nmg-hyper-train}
\centering
\begin{tabular}{ll}
\toprule
\multicolumn{1}{c}{Hyperparameters} & \multicolumn{1}{c}{Value} \\ \midrule
Pre-Training Masking Probability    & 0.05                     \\
Pre-Training Learning Rate          & 0.00002                   \\
Pre-Training Epoch                  & 3                         \\
Sampled Pre-training Dataset Size   & 200                       \\
Pre-Training Batch Size             & Chosen from \{8,16,32\}\\
Maximum sequence length in Pre-Training & 512 (QA) or 256 (TC)  \\ \midrule
Fine-Tuning Learning Rate           & 0.00003 (QA) or 0.00002 (TC)\\
Fine-Tuning Epoch                   & 1 (QA) or 5 (TC)          \\
Maximum Training Set Size           & 1000                      \\
Validation set Size                 & 10000                     \\
Fine-Tuning Batch Size              & Chosen from \{8,16,32\}\\
\bottomrule
\end{tabular}
\end{table*}

\begin{table*}[!ht]
\caption{Hyperparameters for LM Training of Meta-Test.}
\label{table:nmg-hyper-test}
\centering
\begin{tabular}{ll}
\toprule
\multicolumn{1}{c}{Hyperparameters} & \multicolumn{1}{c}{Value} \\ \midrule
Pre-Training Masking Probability    & Chosen from \{0.05, 0.15\} \\
Pre-Training Learning Rate          & 0.00002                     \\
Pre-Training Epoch                  & Chosen from \{1, 3\}      \\
Pre-Training Batch Size             & Chosen from \{12,16,24\}  \\
Maximum sequence length in Pre-Training & 512 (QA) or 256 (TC)    \\ \midrule
Fine-Tuning Learning Rate           & 0.00003 (QA) or 0.00002 (TC)\\
Fine-Tuning Epoch                   & 2 (QA) or 3 (TC)            \\
Fine-Tuning Batch Size              & Chosen from \{12,16,32\}  \\
\bottomrule
\end{tabular}
\end{table*}

\subsection{LM Training in Meta-Train (Inner Loop)}
\label{section:meta-training-detail}
We describe detailed hyperparameters in Table~\ref{table:nmg-hyper-train} for language model (LM) training in the meta-training. 
In the case of pre-processing of pre-training dataset, we use the context of triplets in Question Answering and sentences in Text Classification. Especially for emrQA~\citep{emrQA}, we preprocess it to be same as other QA's formats by removing yes-no and multiple answer type questions. Furthermore, we arbitrarily split a train dataset to a train and a validation set by 8 to 2 since \citet{emrQA} do not provide a separate validation set.
For optimization, we used AdamW optimizer~\citep{AdamW}, with a linear learning rate scheduler. 
Each meta-training is done on the two Titan XP or RTX 2080 Ti GPUs and it costs maximum of 2 days in the case of BERT.
The batch size is adequately selected according to the size of the data and the model.

\subsection{LM Training in Meta-Test}
We describe detailed hyperparameters in Table~\ref{table:nmg-hyper-test} for LM training meta-testing.
In the meta-testing, we use same setting described in Section~\ref{section:meta-training-detail} for pre-processing and optimization.
The batch size is also adequately selected according to the size of the data and the model.

Regarding the pre-training epoch and the masking probability $p$ in meta-testing, we use two distinct settings for the baselines and NMG model.
For the baselines, we train the LM for 1 epoch with $p=0.15$ following a conventional setting.
However, we observed that the fewer masking with a more pre-training epoch is much more beneficial in the meta-training on the task with long contexts since it makes the NMG evaluate actions more precisely.
Therefore, for the NMG model, we train the LM for 3 epochs with $p=0.05$, following the setting of the meta-training.

\subsection{Neural Mask Generator Architecture}
The policy network of the NMG model consists of the single self-attention layer and two linear layers. The self-attention layer follows the configuration of the transformer layer of BERT$_{BASE}$. We omit the linear layers of the original transformer~\cite{Transformer} implementation in our architecture. The hidden size of linear layers is 128 and gelu~\cite{GELU} is used as an activation function.
The value network also consists of the same linear layers after mean-pooling of word representations.
The total number of parameters of the NMG model is approximately 2.5M, which is far smaller than the conventional language model.

\subsection{Hyperparameter Searching}
For searching proper hyperparameter, we use a manual tuning which tries conventional hyperparameters for the reinforcement learning (RL) and the language model (LM) training. A selection criterion depends on the result of the meta-testing. Especially, we set the criterion to F1 score and accuracy for question answering and classification task respectively.

\section{More Examples}
\label{sec:examples}
To show the masking strategy from our NMG model, we additionally append additional examples from various datasets used in our experiments.

\begin{figure*}[ht!]
    \centering
    \includegraphics[width=1.0\linewidth]{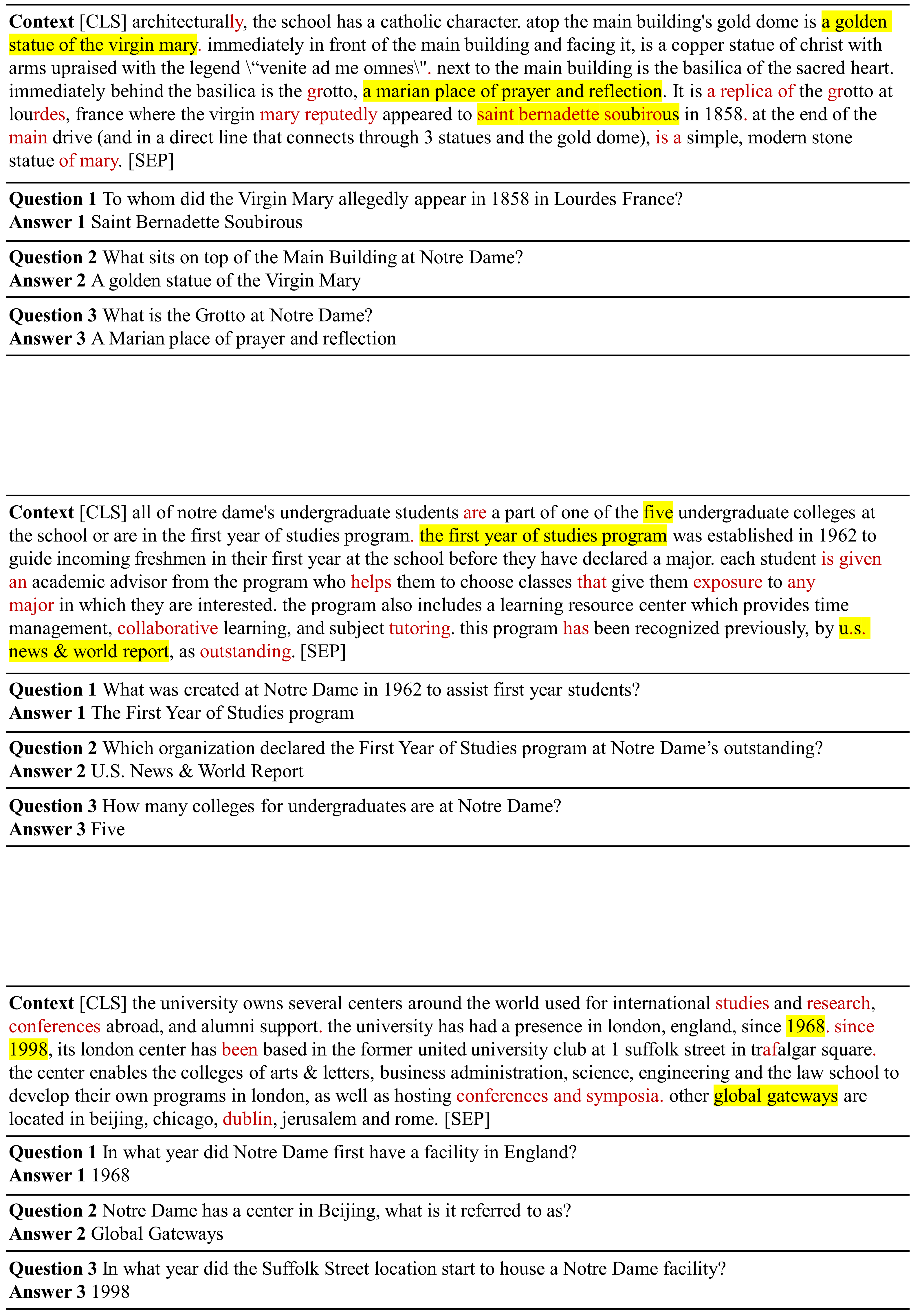}
    \caption{\textbf{SQuAD} Examples of masked tokens using the Neural Mask Generator (NMG). The red mark and yellow box indicates masked tokens by the model and the answer given a question, respectively.}
    \vspace{-0.2cm}
\end{figure*}

\begin{figure*}[ht!]
    \centering
    \includegraphics[width=1.0\linewidth]{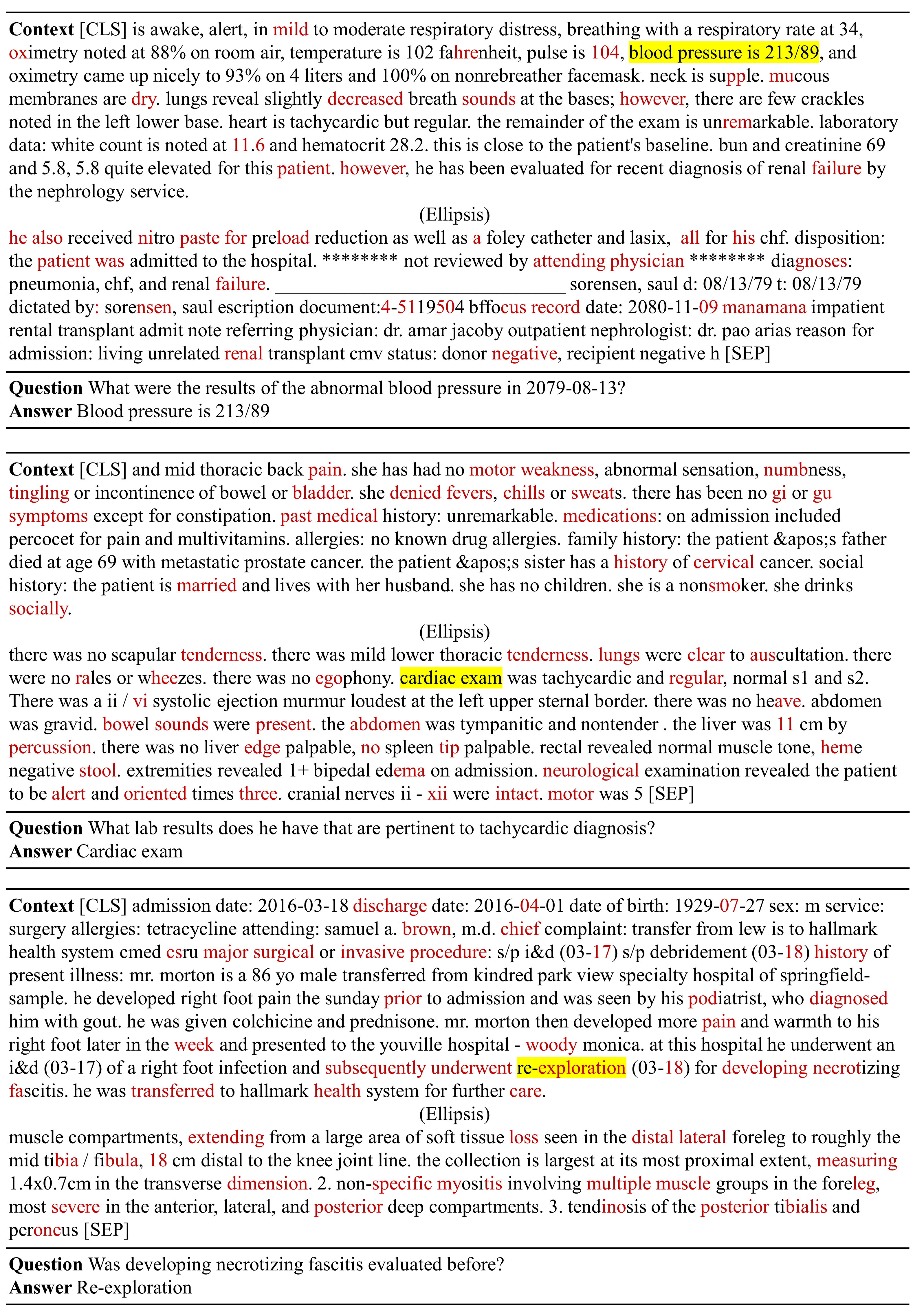}
    \caption{\textbf{emrQA} Examples of masked tokens using the Neural Mask Generator (NMG). The red mark and yellow box indicates masked tokens by the model and the answer given a question, respectively.}
    \vspace{-0.2cm}
\end{figure*}

\begin{figure*}[ht!]
    \centering
    \includegraphics[width=1.0\linewidth]{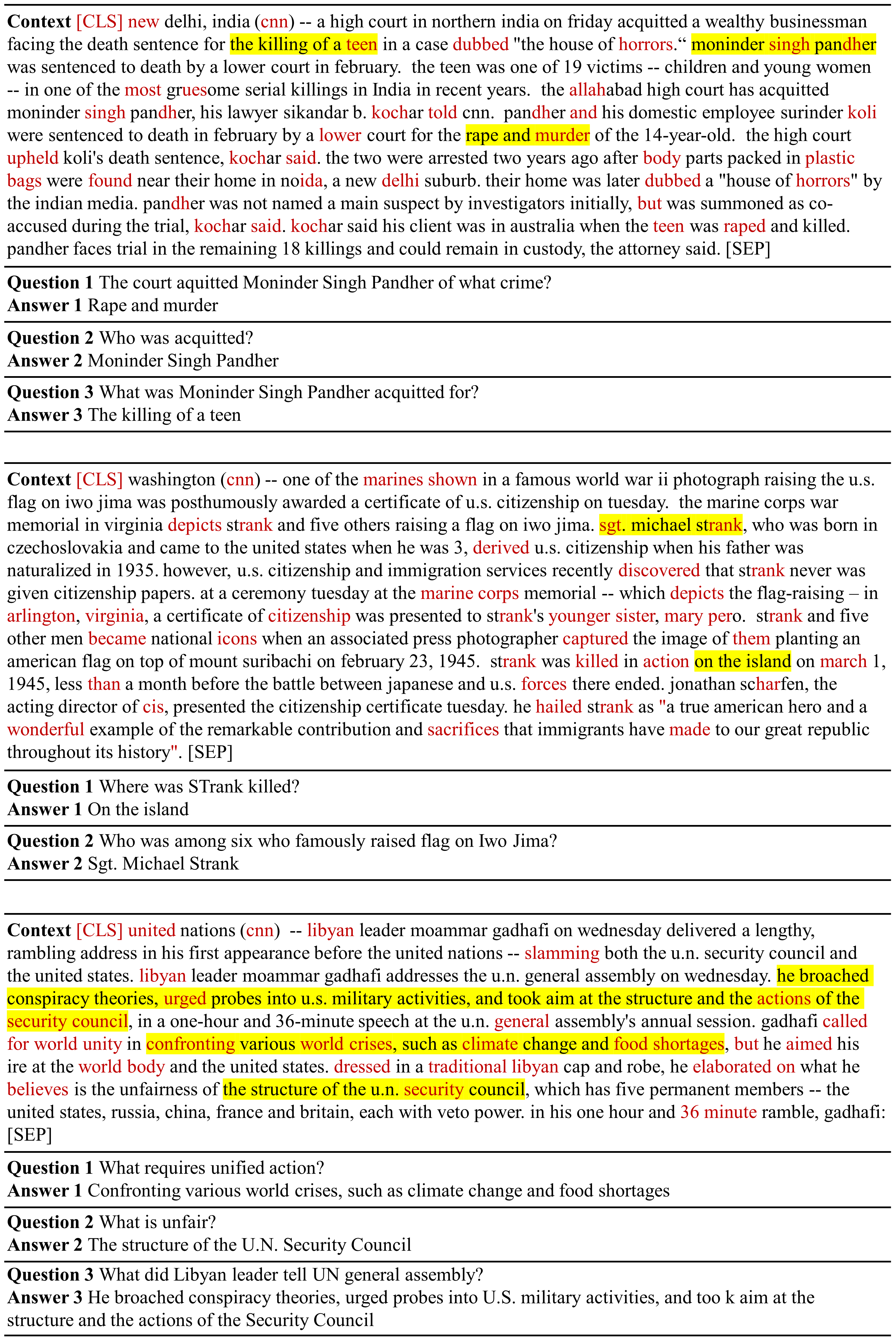}
    \caption{\textbf{NewsQA} Examples of masked tokens using the Neural Mask Generator (NMG). The red mark and yellow box indicates masked tokens by the model and the answer given a question, respectively.}
    \vspace{-0.2cm}
\end{figure*}

\begin{figure*}[ht!]
    \centering
    \includegraphics[width=1.0\linewidth]{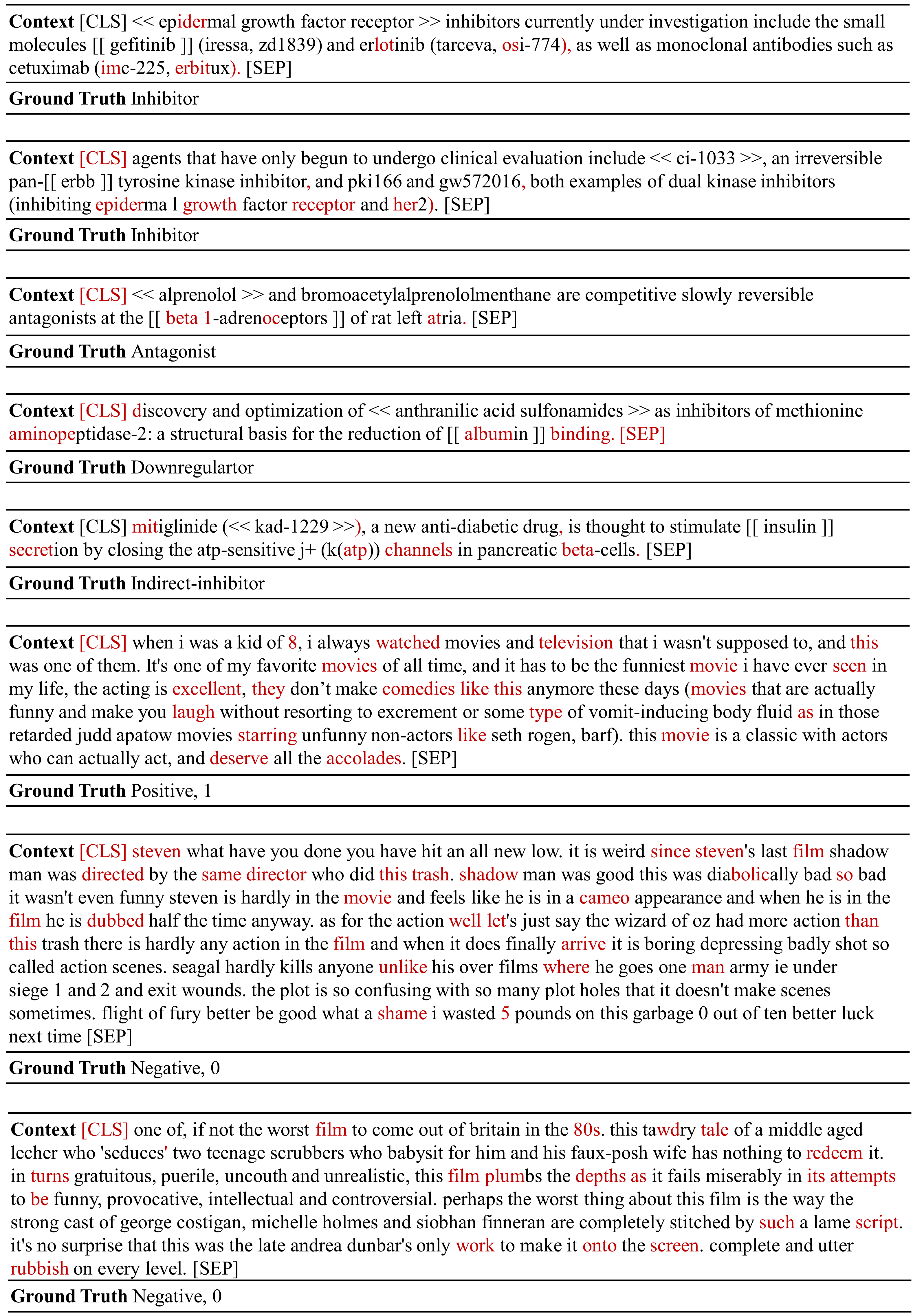}
    \caption{\textbf{ChemProt and IMDb} Examples of masked tokens using the Neural Mask Generator (NMG). The red mark indicates masked tokens by the model.}
    \vspace{-0.2cm}
\end{figure*}

\end{document}